\pdfoutput=1

\documentclass[11pt]{article}

\usepackage[final]{acl}

\usepackage{times}
\usepackage{latexsym}

\usepackage[T1]{fontenc}

\usepackage[utf8]{inputenc}

\usepackage{microtype}

\usepackage{inconsolata}

\usepackage{graphicx}
\usepackage{booktabs}
\usepackage{pdfcomment}
\usepackage{xcolor}

\newcommand{\repl}[1]{#1}

\title{HausaNLP at SemEval-2025 Task 11: Hausa Text Emotion Detection}

\author{%
  Sani Abdullahi Sani$^{1,2}$, Salim Abubakar$^{1,3}$, Falalu Ibrahim Lawan$^{1,4}$, \\
  \bf Abdulhamid Abubakar$^1$, Maryam Bala$^1$\\
  \small $^1$HausaNLP, $^2$University of the Witwatersrand,  $^3$Federal Polytechnic Daura, $^4$Kaduna State University \\
  \footnotesize{\texttt{\textbf{correspondence:} saniabdullahisani1@students.wits.ac.za}}
}

\begin{document}
\maketitle
\begin{abstract}
This paper presents our approach to multi-label emotion detection in Hausa, a low-resource African language, for SemEval Track A. We fine-tuned AfriBERTa, a transformer-based model pre-trained on African languages, to classify Hausa text into six emotions: anger, disgust, fear, joy, sadness, and surprise. Our methodology involved data preprocessing, tokenization, and model fine-tuning using the Hugging Face Trainer API. The system achieved a validation accuracy of 74.00\%, with an F1-score of 73.50\%, demonstrating the effectiveness of transformer-based models for emotion detection in low-resource languages.
\end{abstract}


\section{Introduction}

Emotion detection in text is a fundamental Natural Language Processing (NLP) task with far-reaching applications in sentiment analysis, social media monitoring, and mental health support. While significant progress has been made in high-resource languages, low-resource languages like Hausa remain underrepresented in the NLP landscape. \repl{A} key \repl{challenge} is the limited availability of annotated Hausa datasets for emotion detection, \repl{hindering} the development of robust models. \repl{We} \repl{address} this gap by leveraging \repl{the} newly available BRIGHTER \repl{dataset} \cite{muhammad2025brighterbridginggaphumanannotated} and fine-tuning a transformer-based model for multi-label emotion detection in Hausa.

The scarcity of comprehensive emotion lexicons and annotated corpora \repl{complicates} \repl{the} \repl{development} and \repl{evaluation of} emotion detection systems for low-resource languages \citep{kabir-etal-2023-bemolexbert, Al-Wesabi2023, raychawdhary-etal-2023-seals}. \repl{Furthermore}, these languages exhibit rich morphological variations, syntax, and semantic differences that are not well-captured by models trained on high-resource languages \citep{Marreddy2022, Girija2023}. \repl{Code}-mixing, where multiple languages are used within the same text, \repl{further} \repl{complicates} emotion detection in linguistically diverse contexts \citep{raychawdhary-etal-2023-seals, sonu-etal-2022-identifying}.

\repl{We} \repl{leverage} AfriBERTa, a transformer model specifically trained on African languages \repl{and} fine-tuned for multi-class emotion classification. \repl{We preprocessed and tokenized the} text data using the AfriBERTa tokenizer, and \repl{then} fine-tuned \repl{the model} on the BRIGHTER Hausa dataset. This work contributes to emotion detection \repl{research} in low-resource languages by demonstrating effective methods for adapting transformer models to Hausa. Our findings highlight the potential of leveraging pre-trained models like AfriBERTa \citep{ogueji-etal-2021-small} for emotion detection tasks in low-resource African languages.

\section{Background}

\subsection{SemEval Task Overview}
SemEval-2024 Track A, the Multi-label Emotion Detection task under Task 11: Bridging the Gap in Text-Based Emotion Detection \citep{muhammad-etal-2025-semeval}, focuses on classifying text from multiple languages into six emotion categories: anger, disgust, fear, joy, sadness, and surprise. While the task encompasses a wide range of languages, our team (HausaNLP) \repl{focused} on Hausa, a Chadic language spoken across West and Central Africa by approximately 88 million people, including 54 million native speakers and 34 million second-language users primarily in Nigeria, Niger, and neighboring countries \citep{Wolff2024, Eberhard2024}. The \repl{system} input is a \repl{Hausa} text sample, and the output is a one-hot encoded vector indicating the presence or absence of each emotion. This task is particularly challenging due to the nuances of emotion expression in Hausa, as well as the limited availability of annotated data for low-resource languages. 

\subsection{Hausa Dataset}

\begin{figure}[t]
  \centering
  \includegraphics[width=\linewidth, clip, trim=8 25 15 20]{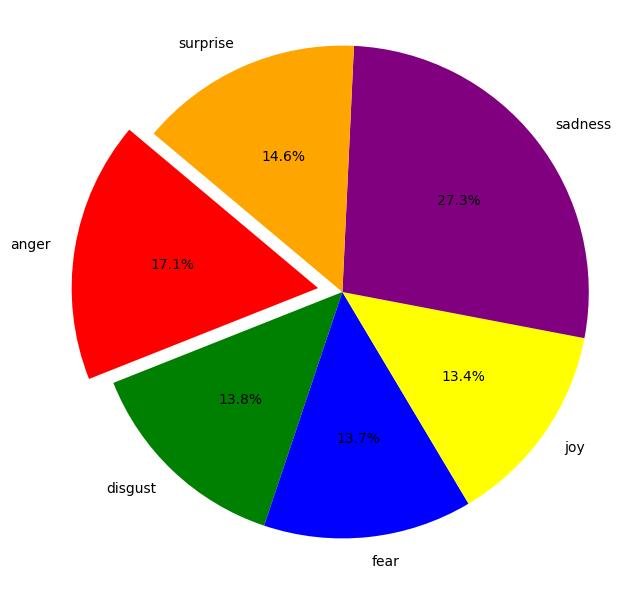}
  \caption{Distribution of emotions in the Hausa text dataset. The bar plot shows the frequency of each emotion category (joy, sadness, anger, fear, surprise, and disgust) present in the dataset, obtained through manual annotation by three independent annotators}.
  \label{hausadataset}
\end{figure}

The Hausa dataset is divided into \repl{train} (approximately 2,145 samples), validation (356 samples), and test (1,080 samples\repl{) splits}. The complete distribution of the dataset is shown in Figure ~\ref{hausadataset}. Each sample is annotated with one or more emotion labels, represented as a one-hot encoded vector as shown in ~\ref{tab:sample_dataset}. The dataset exhibits class imbalance, with emotions \repl{like} joy and sadness being more prevalent than fear and disgust. This imbalance posed a challenge during model training and evaluation.

\section{Related Work on Low-Resource Emotion Detection}



\repl{Researchers} have proposed \repl{several} approaches to address the challenges of emotion detection in low-resource settings. \repl{The} AfriSenti-SemEval 2023 Shared Task provided annotated datasets for \repl{low-resource} languages like Hausa and Igbo \citep{raychawdhary-etal-2023-seals}. Techniques such as few-shot and zero-shot learning have been employed to augment datasets and improve model performance in data-scarce environments \citep{Agarwal2025105, hasan-etal-2024-zero}.

Pre-trained models like mBERT, \repl{Cross-Lingual Model - RoBERTa (}XLM-R), and BanglaBERT have been fine-tuned for specific low-resource languages, showing improved performance in emotion detection tasks \citep{Raychawdhary2023, kabir-etal-2023-bemolexbert, Raychawdhary2024}. Hybrid models combining lexicon features with transformers have also been explored to enhance emotion classification \citep{kabir-etal-2023-bemolexbert}. \repl{Deep} learning models (e.g., \repl{Convolutional Neural Networks (}CNNs), \repl{Deep Neural Networks (}DNNs)) and hybrid approaches (e.g., TCSO-DGNN) have been proposed to better capture the emotional nuances in text \citep{Jadon20231329, Bao2025}.

AfriBERTa, a transformer model pre-trained on a diverse corpus of African languages, has shown promising results for tasks like text classification and named entity recognition in Hausa and other African languages \citep{ogueji-etal-2021-small}. Our work builds on these advancements by fine-tuning AfriBERTa for emotion detection in Hausa, addressing the unique challenges of low-resource language processing.

Future directions include multimodal emotion detection, \repl{incorporating} additional data modalities (e.g., audio, video) to complement textual analysis \citep{Sarbazi-Azad2021465, Wang2024}, developing cross-lingual and multilingual models that generalize across multiple low-resource languages \citep{Raychawdhary2023, Raychawdhary2024}, and addressing ethical considerations related to data privacy, bias, and deployment of emotion detection systems in diverse cultural contexts \citep{Agarwal2025105}.

\begin{table*}
  \centering
  \caption{Sample \repl{entries} from the Hausa \repl{emotion} \repl{detection} \repl{dataset. The table displays five example entries from the dataset, including the Hausa text, its English translation, and the corresponding emotion labels assigned by human annotators}.}
  \label{tab:sample_dataset}
  
  \begin{tabular}{lp{6cm}cccccc}
    \toprule
    \textbf{ID} & \textbf{Text} & \textbf{Anger} & \textbf{Disgust} & \textbf{Fear} & \textbf{Joy} & \textbf{Sadness} & \textbf{Surprise} \\
    \midrule
    1 & Kotu Ta Yi Hukunci Kan Shari'ar Zaben Dan Majalisar PDP, Ta Yi Hukuncin Bazata & 0 & 0 & 0 & 0 & 0 & 1 \\
    \midrule
    2 & Toh fah inji 'yan magana suka ce ``ana wata ga wata'' & 0 & 0 & 0 & 0 & 0 & 1 \\
    \midrule
    3 & Bincike ya nuna yan Najeriya sun fi damuwa da rashin tsaro da talauci fiye da korona & 0 & 0 & 1 & 0 & 1 & 0 \\
    \bottomrule
  \end{tabular}
\end{table*}

\section{Methodology}

\begin{figure}[t]
  \centering
  \includegraphics[width=0.8\linewidth, clip, trim=0 0 5 0]{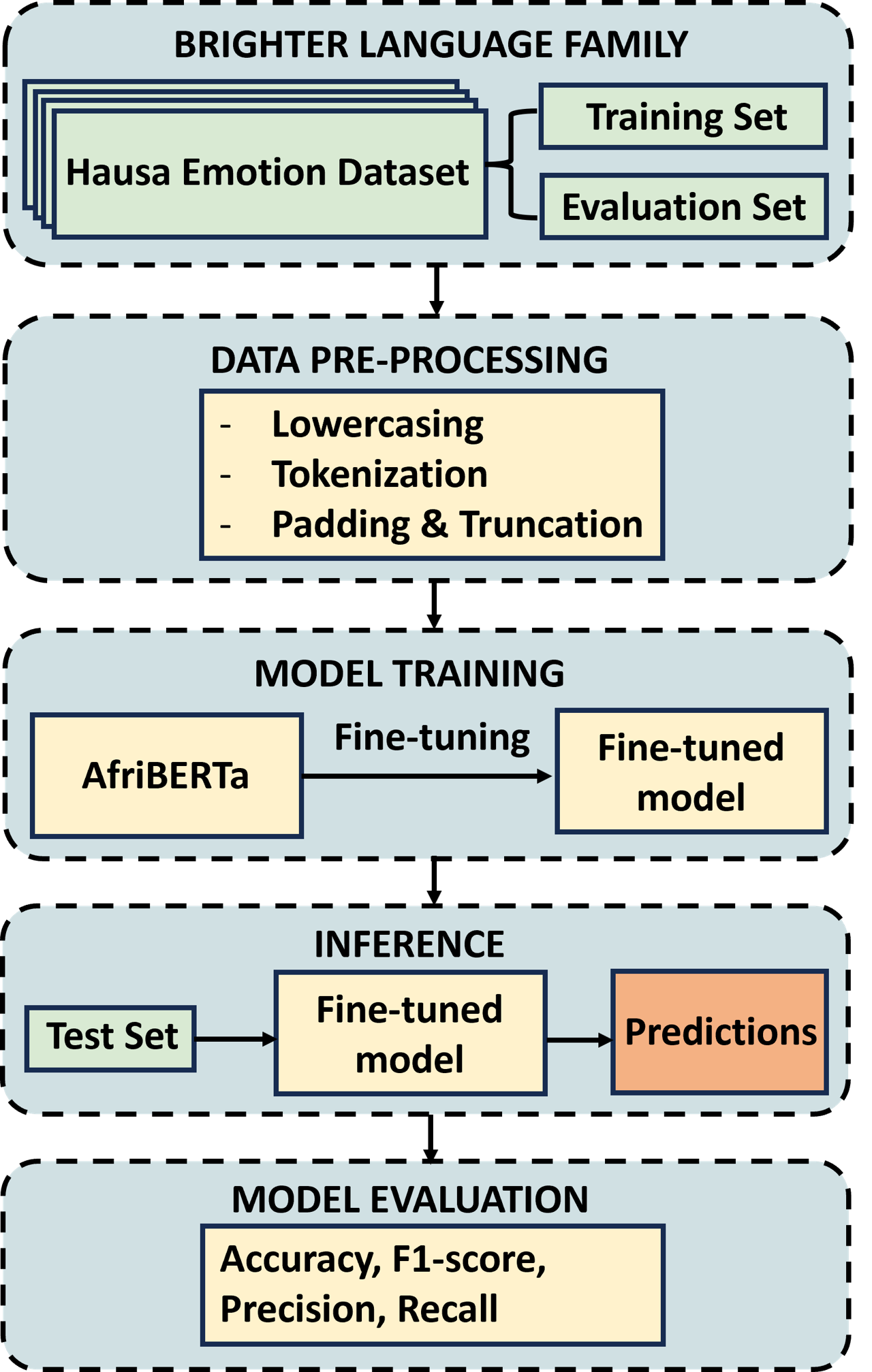}
  \caption{Multi-label \repl{emotion} \repl{detection} \repl{system overview. The diagram illustrates the complete multi}-\repl{label} \repl{emotion} \repl{detection process, which includes data preprocessing, feature extraction using TF-IDF and word embeddings, classification with a BiLSTM network, and evaluation using F1-score}}
  \label{system overview}
\end{figure}

\subsection{Data Preprocessing} 

To prepare the data for training, a preprocessing pipeline was implemented. First, the text data was cleaned by converting it to lowercase and stripping extra spaces. This step ensured consistency and improved tokenization. Next, \repl{we mapped} the one-hot encoded labels to a single integer label representing the dominant emotion. This simplified the classification task into a multi-class problem with six classes.

\subsection{Tokenization and Formatting} 

\repl{We tokenized the} preprocessed text data using the AfriBERTa tokenizer, a pre-trained tokenizer designed for African languages, including Hausa. The tokenizer was configured to truncate sequences longer than 128 tokens and pad shorter sequences to ensure uniform input sizes. The tokenized dataset was then formatted into PyTorch tensors, containing the input\_ids, attention\_mask, and label columns. This prepared the data for the transformer model.

\subsection{Model Selection and Fine-Tuning} 

\repl{We selected the} AfriBERTa Small model, a compact version of the AfriBERTa architecture, \repl{primarily} \repl{due} \repl{to} compute \repl{limitations}. \repl{We} fine-tuned \repl{the model} for sequence classification with six output labels corresponding to the six emotions.

The fine-tuning process was conducted using the Hugging Face Trainer \repl{Application Programming Interface (}API). The training arguments were configured with a learning rate of 2e-5, a batch size of 8, and 5 epochs. To optimize training, \repl{we enabled} mixed precision (fp16) \repl{and} \repl{applied} a warmup schedule with 500 steps to gradually increase the learning rate at the start of training. The model was evaluated on the validation set after each epoch, and the best model was saved based on validation accuracy.

\subsection{Evaluation Metrics} 

\repl{We} \repl{evaluated} the model's performance \repl{during} \repl{training} using standard classification metrics: accuracy, precision, recall, and F1-score. These metrics were computed using the classification\_report function from the sklearn.metrics library. \repl{Evaluation} was performed on the validation set after each epoch to monitor the model's progress and ensure \repl{effective} learning.

\subsection{Saving and Inference} 
After fine-tuning, the best-performing model was saved to disk for future use. The model and tokenizer were saved. For inference, \repl{we loaded} the model using the Hugging Face pipeline API, \repl{simplifying} predictions on new text samples. The model was tested on a sample Hausa text, and the predicted emotion was displayed.

\subsection{Predictions on Test Set} 
To evaluate the model's performance on unseen data and submit to SemEval 2025 Task 11, \repl{we} made \repl{predictions} on the test set. \repl{We loaded the} test set from a \repl{Comma-Separated Values (}CSV) file, and the model predicted the dominant emotion for each text sample. \repl{We converted the} predictions into one-hot encoded labels and saved \repl{them} to a new CSV file. This file contained the original text samples along with the predicted emotion labels, allowing for further analysis and evaluation.

\section{Results and Discussion}

\begin{table*}[h!]
\centering
\caption{Training and \repl{validation} \repl{metrics} for the \repl{fine}-\repl{tuned} AfriBERTa model. The graph presents the accuracy and F1-score curves for both the training and validation sets across all training epochs, demonstrating the model's learning progress and generalization capability}.
\label{tab:results}
\renewcommand{\arraystretch}{1.2}
\begin{tabular}{lcccc}
\toprule
& \textbf{Accuracy} & \textbf{Precision} & \textbf{Recall} & \textbf{F1-Score} \\ \midrule
\textbf{Train} & 0.7416 ± 0.02     & 0.7515 ± 0.02      & 0.7348 ± 0.02   & 0.7393 ± 0.02     \\ \midrule
\textbf{Val}   & 0.7400 ± 0.02     & 0.7500 ± 0.02      & 0.7300 ± 0.02   & 0.7350 ± 0.02     \\ \bottomrule
\end{tabular}
\end{table*}

The fine-tuned AfriBERTa model demonstrated consistent improvement over the training epochs, as shown in Table~\ref{tab:results}. The model achieved its best performance in the fifth epoch, with a training accuracy of \textbf{74.16\%} and an F1-score of \textbf{73.93\%}. Precision and recall scores were also strong, at \textbf{75.15\%} and \textbf{73.48\%}, respectively. These results indicate that the model effectively learned to classify emotions in Hausa text, despite the challenges posed by class imbalance and noisy data.

\begin{figure}[t]
  \centering
  \includegraphics[width=\linewidth, clip, trim=10 10 5 0]{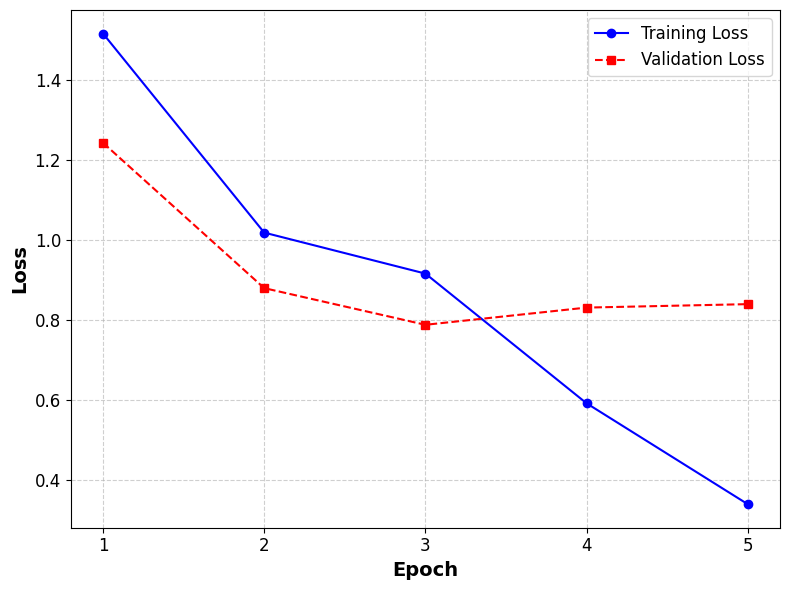}
  \caption{Comparison of Training and Validation Loss Across Epochs: The plot illustrates the progression of training and validation loss over five epochs. The training loss steadily decreases, indicating that the model is learning from the data. However, the validation loss initially decreases but later fluctuates, suggesting potential overfitting or variations in generalization performance.}
\end{figure}

Additionally, the model showed steady progress, with the training loss decreasing from \textbf{1.5168} in the first epoch to \textbf{0.3385} in the fifth epoch. The validation loss also decreased initially but stabilized around \textbf{0.8393} in the final epoch, suggesting that the model reached a point of convergence. The model's performance on the validation set was consistent with the training results, achieving an accuracy of \textbf{74.00\%} and an F1-score of \textbf{73.50\%}, demonstrating its ability to generalize to unseen data.

The model performed particularly well on emotions like \textbf{joy} and \textbf{sadness}, which were more prevalent in the dataset. However, it struggled slightly with underrepresented emotions such as \textbf{fear} and \textbf{disgust}, likely due to their limited representation in the training data. This highlights the importance of addressing class imbalance in future work, potentially through techniques like data augmentation or weighted loss functions. Overall, the results underscore the effectiveness of AfriBERTa for emotion detection in low-resource languages and provide a strong baseline for future research in this domain.

\section{Conclusion}

In this work, we fine-tuned AfriBERTa for multi-label emotion detection in Hausa text, achieving an accuracy of 74.00\% and an F1-score of 73.50\%. The model performed well on prevalent emotions like joy and sadness but struggled with underrepresented emotions such as fear and disgust, highlighting the challenge of class imbalance. Our approach demonstrates the effectiveness of transformer-based models for low-resource language tasks. Future work could address class imbalance through data augmentation or weighted loss functions and extend this work to other African languages. This study provides a strong baseline for emotion detection in Hausa and contributes to the development of inclusive NLP systems.

\section{Limitations}
One major limitation of this work is the variability in Hausa dialects. The dataset primarily represents standard Hausa, which may not generalize well to regional dialects or informal variations commonly used in social media and conversational settings. This could lead to misclassification of emotions when processing text from underrepresented dialects.  

Another limitation is the generality of the dataset. While the dataset is carefully curated, it may not fully capture the diversity of emotions expressed in different contexts, such as sarcasm, code-mixing with English or Arabic, and cultural-specific expressions. This limits the robustness of the model when applied to unseen, real-world data.  

Additionally, computational constraints influenced the choice of AfriBERTa-small instead of larger transformer models. While this ensures efficiency, it may come at the cost of lower accuracy compared to models with higher capacity. The trade-off between computational efficiency and model performance is a key consideration for practical deployment.  

Lastly, deep learning models often lack interpretability, making it difficult to explain why a particular emotion was predicted. This could pose challenges in high-stakes applications, such as mental health monitoring or crisis detection, where transparency is crucial.



\section*{Acknowledgements}
The authors acknowledge Google DeepMind for a study grant.

\bibliography{anthology,custom}

\appendix

\label{sec:appendix}


\end{document}